\documentclass[runningheads]{llncs}
\usepackage{graphicx}
\usepackage{hyperref}

\usepackage{caption}
\usepackage{subcaption}

\begin{document}
\title{Avalanche RL: a Continual Reinforcement Learning Library}
%
%
\author{Nicolò Lucchesi \and
Antonio Carta
\and
Vincenzo Lomonaco \and Davide Bacciu
}
\authorrunning{N. Lucchesi et al.}
%
\institute{Deparment of Computer Science, University of Pisa\\
\and
\email{nicolo.lucchesi@gmail.com}
\and
\email{antonio.carta@di.unipi.it}\\
\and
\email{vincenzo.lomonaco@unipi.it}
\and
\email{bacciu@unipi.it}}
%
\maketitle              
\begin{abstract}
Continual Reinforcement Learning (CRL) is a challenging setting where an agent learns to interact with an environment that is constantly changing over time (the stream of \textit{experiences}).
In this paper, we describe \texttt{Avalanche RL}, a library for Continual Reinforcement Learning which allows users to easily train agents on a continuous stream of tasks. \texttt{Avalanche RL} is based on PyTorch \cite{pytorch} and supports any OpenAI \texttt{Gym} \cite{gym} environment. Its design is based on Avalanche \cite{avalanche}, one of the most popular continual learning libraries, which allow us to reuse a large number of continual learning strategies and improve the interaction between reinforcement learning and continual learning researchers.
Additionally, we propose Continual Habitat-Lab, a novel benchmark and a high-level library which enables the usage of the photorealistic simulator Habitat-Sim \cite{habitat} for CRL research.
Overall, \texttt{Avalanche RL} attempts to unify under a common framework continual reinforcement learning applications, which we hope will foster the growth of the field.

\end{abstract}
\section{Introduction}
\begin{figure}
    \centering
    \includegraphics[width=0.7     \textwidth]{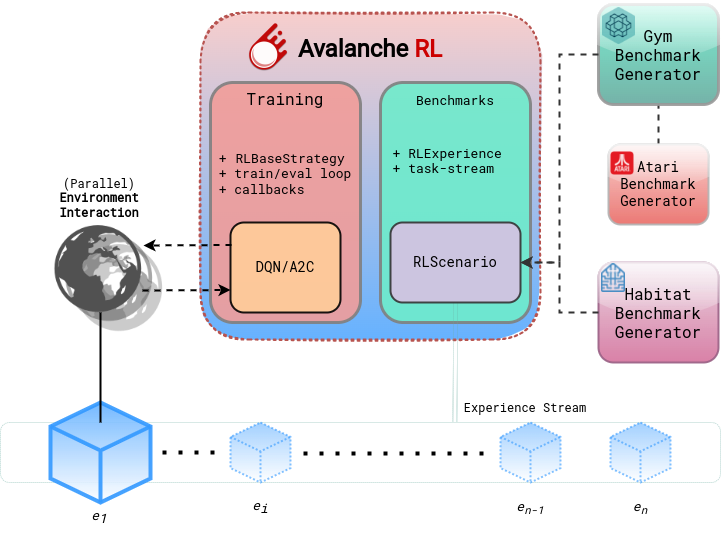}
    \caption{\texttt{Avalanche RL} core-functionalities overview. The Benchmarks module capabilities, providing access to a stream of environments, are addressed in Section \ref{benchmarks}. Data is obtained through (parallel, Section \ref{vec-env}) interaction with the stream and it is consumed by the algorithm in the learning process, as motivated in Section \ref{avlrl-train}. Streams can be easily created through benchmark generators (right-hand side). }
    \label{fig:diagram}
\end{figure}

Recent advances in data-driven algorithms, the so-called Deep Learning revolution, has shown the possibility for AI algorithms to achieve unprecedented performances on a narrow set of specific tasks. 
On the contrary, humans are able to quickly learn new tasks and generalize to novel scenarios. Continual Learning (CL) in the same way seeks to develop
data-driven algorithms able to incrementally learn behaviors from a stream of data.
Reinforcement Learning (RL) is yet another Machine Learning paradigm which formulates the
learning process as a sequence of interactions between an agent and the environment. The agent must learn off of this interaction how to achieve a goal of a particular task by taking actions in the environment while receiving a (scalar) reward.
Continual Reinforcement Learning (CRL) combines the non-stationarity assumption of a stream
of data with the RL setting, having an agent learn multiple tasks in sequence.\\\\
While still in its early stages, CRL has seen a rising interest in publications in recent years (according to Dimensions \cite{dimensions} data). To support this growth, we focus on benchmarks and tools, introducing AvalancheRL: we extend Avalanche \cite{avalanche}, the staple framework for Continual or Lifelong Learning, to support Reinforcement Learning in order to seamlessly train agents on a continuous stream tasks.\\\\
Existing RL libraries \cite{sb3,ray-rl,salina,kerasrl} do not focus on lifelong applications and force users to write custom code to develop continual solutions. Avalanche gives us re-usability by providing pre-implemented CL strategies as well as code structure when experimenting with them, but lacked support altogether when coming to RL. Related CRL projects instead either focus on providing a specific benchmark \cite{continual-world} or combine multiple frameworks results \cite{sequoia}, limiting the overall flexibility and methods customization options.\\
\texttt{Avalanche RL} attempts to address both problems aiming to offer a malleable framework encompassing a variety of RL algorithms with fine-grained control over their internals, leveraging pre-existing CL techniques to learn efficiently from the interaction with multiple environments. 
In particular, we support any environment exposing the OpenAI Gym \texttt{gym.Env} interface.\\\\
The availability of compelling benchmarks has always lead the progress of data-driven algorithms \cite{mnist,cifar10,imagenet}, therefore our second effort is aimed at providing a challenging dataset for realistic Continual Reinforcement Learning.\\
Habitat-Lab allows an embodied agent to roam a photorealistic (typically indoor) scene in the attempt of solving a particular task; unfortunately, it does not offer support for the continual scenario. 
Therefore, we developed Continual Habitat-Lab, a high-level library enabling the usage of Habitat-Sim \cite{habitat} for CRL, allowing the creation of sequences of tasks while integrating with \texttt{Avalanche RL}.\\\\
We first outline the design principles that guided the development of \texttt{Avalanche RL} (Section \ref{principles}), describe its structure (Figure \ref{fig:diagram}) and go over the main features of the framework with code examples (Section \ref{avrl}). We then introduce Continual Habitat-Lab and describe its integration with \texttt{Avalanche RL} (Section \ref{chl}).\\\\
All the source code of the work hereby presented is publicly available on GitHub for both \texttt{Avalanche RL}\footnote{\url{https://github.com/continualAI/avalanche-rl}} and Continual Habitat-Lab\footnote{\url{https://github.com/NickLucche/continual-habitat-lab}}. 

\section{Design Principles} \label{principles}
\texttt{Avalanche RL} is built as an extension of Avalanche \cite{avalanche}, and it retains the same design principles and a similar API. The target users are practitioners and researchers, and therefore the library must be simple, allowing to setup an experiment with a few lines of code, as well as highly customizable. As a result, \texttt{Avalanche RL} provides high-level APIs with ready-to-use components, as well as low-level features that allow heavy customization of existing implementations by leveraging an exhaustive callback system (Section \ref{avlrl-train}).\\
\texttt{Avalanche RL} codebase is comprises 5 main modules: \textbf{Benchmarks}, \textbf{Training}, \textbf{Evaluation}, \textbf{Models}, and \textbf{Logging}. We give a brief overview of them in the remainder of this section, but we refer the reader to~ \cite{avalanche} for more details about the general architecture of \texttt{Avalanche}.

\subsubsection{Benchmarks} maintains a uniform API for data handling, generating a \textit{stream} of data from one or more datasets, conveniently divided into temporal \textit{experiences}; this is the core abstraction over the \textit{task stream} formalism which is distinctive of CL and it is accessible through a \textbf{Scenario} object. In order to create benchmarks more easily, this module provides \textit{benchmark generators} which allow one to specify particular configurations through a simple API.  

\subsubsection{Training} provides all the necessary utilities concerning model training. It includes simple and efficient ways of implementing new \textit{strategies} as well as a set pre-implemented CL baselines and state-of-the-art algorithms. A \textbf{Strategy} abstracts a general learning algorithm implementing a training and an evaluation loop while consuming experiences from a \textit{benchmark}. Continual behaviors can be added when needed through \textbf{Plugins}: they operate latching on the callback system defined by Strategies and are designed in such a modular way so that they can be easily composed to provide hybrid behaviors. 
    
\subsubsection{Evaluation} provides all the utilities and metrics that can help evaluate a CL algorithm. Here we can find \textit{pluggable} metric monitors such as (Train/Test/Batch) Accuracy, RAM, CPU and GPU usage, all designed with the same modularity principles in mind.

\subsubsection{Models} contains several model architectures and pre-trained models that can be used for continual learning experiments (similar to \texttt{torchvision.models}), from simple customizable networks to implementation of state-of-the-art models.
\subsubsection{Logging} includes advanced logging and plotting features with the purpose of visualizing the metrics of the Evaluation module, such as highly readable output, file and \texttt{TensorBoard} support.

\subsection{Notation}

We adopt the well renowned notation from  \cite{sutton-barto} for Reinforcement Learning related formulations while we make use of the formalization introduced in  \cite{cl-robotics} regarding Continual Learning.\\
In particular, we refer to the RL problem as consisting of a tuple of five elements commonly denoted as $<\mathcal{S}, \mathcal{A}, \mathcal{R}, \mathcal{P}, \gamma>$ in the MDP formulation, where $\mathcal{S}$ and $\mathcal{A}$ are sets of \textbf{states} and \textbf{actions}, respectively. $\mathcal{R}$ or $r()$ is the \textbf{reward function}, with $r(s, a, s')$ being the expected immediate reward for transition from state $s\in\mathcal{S}$ to $s'\in\mathcal{S}$ under action $a\in\mathcal{A}$. $\mathcal{P}$ or $p()$ is the \textbf{transition function} defining the dynamics of the environment, with $p(s', r | s, a)$ denoting the probability of transitioning from $s$ into $s'$ with scalar reward $r$ under $a$. Finally, $\gamma$ represents the discount factor which weights the importance of immediate and future rewards.\\
An agent follows a policy $\pi$, which maps states to action probabilities. In Deep RL, learned policies are parameterized function (such as a neural network) which we indicate with $\pi_\theta$.\\  
We refer to a \textit{Dataset} as a collection of samples $\{x_i\}_i^N$, optionally with labels $\{<x_i, y_i>\}_i^N$ in the case of supervised learning. We then denote a general task to be solved by some agent with $\tau$ and define the data relative to that task with $D_\tau$.

\section{Avalanche RL} \label{avrl}
CRL applications in \texttt{Avalanche RL} are implemented by modeling the interaction between core components: the task-stream abstraction (i.e., the continuously changing environment) and the RL strategy (i.e., the agent and its learning algorithm).

\texttt{Avalanche RL} implements these two components in the \textbf{Benchmarks} and \textbf{Training} module, respectively. In the remainder of this section, we describe the environment and the implementation of its continual shift in Section \ref{benchmarks}. Then, in Section \ref{avlrl-train}, we describe the implementation of RL algorithms and their integration in the Training module. Section \ref{vec-env} and \ref{extras} highlight some important implementation details and useful features offered by the framework, such as the automatic parallelization of the RL environment.

\subsection{Benchmarks: Stream of Environments} \label{benchmarks}

Most continual learning frameworks  \cite{cl-robotics} assume that the stream of data is made of static datasets of a fixed size. Instead, in CRL problems the stream consists of different environments, and samples are obtained through the interaction between the agent and the environment.

To support streams of environments, \texttt{Avalanche RL} defines a stream $S = \{e_1, e_2, ..\}$ as a sequence of experiences $e_i$, where each experience provides access to an environment with which the agent can interact to generate state transitions (samples) online. Over time, this means that the agent learns by interacting with a stream of environments $\{\mathcal{E}_1, \mathcal{E}_2, ..\}$, as in Figure \ref{fig:diagram}. In the source code, \texttt{RLExperience} is the class which defines the CRL experience. 

Using this \textbf{task-stream abstraction}, it is easy to define CRL benchmarks as a set of parallel streams of environments. Notice that each experience may be a small shift, such as a change in the background, as well as a completely different tasks, such as a different game. Different tasks may provide a task label which can be used by the agent to distinguish among them. The \texttt{RLScenario} is the class responsible for the CRL benchmark's definition, and it can be thought as a container of streams.\\

\begin{figure}
\centering
\begin{subfigure}{.5\textwidth}
  \centering
  \includegraphics[width=1.\linewidth]{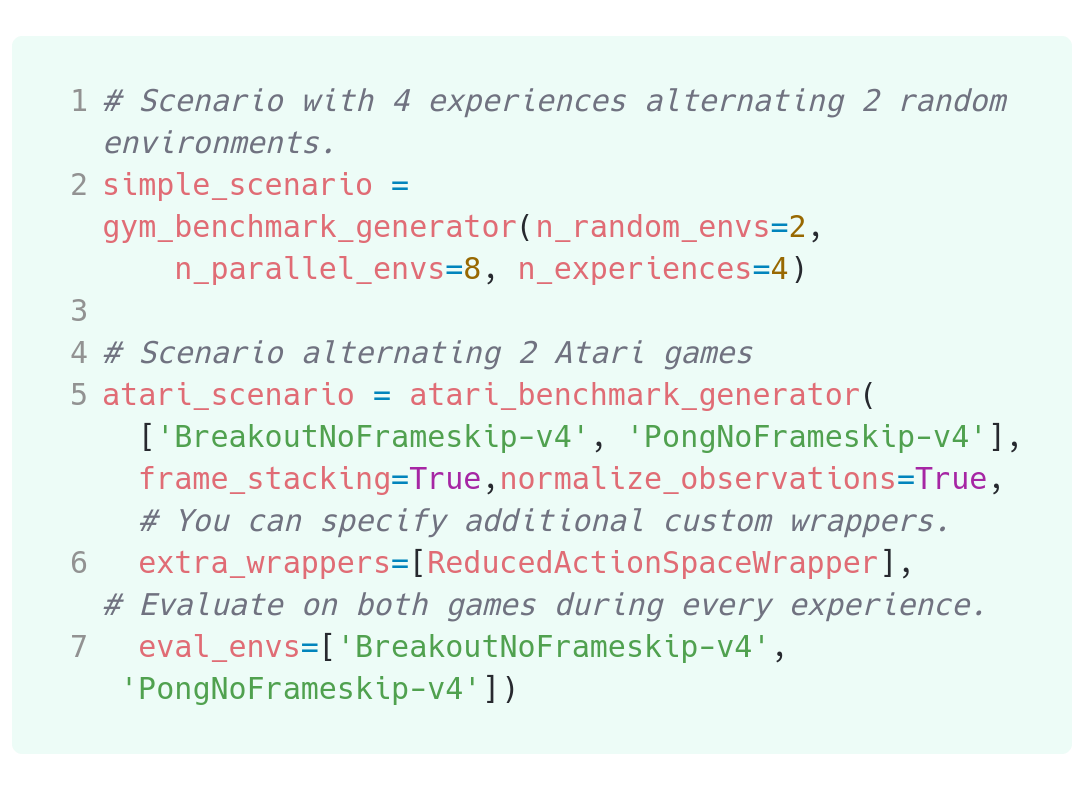}
  \caption{Benchmark creation}
  \label{fig:sub1}
\end{subfigure}%
\begin{subfigure}{.5\textwidth}
  \centering
  \includegraphics[width=1.\linewidth]{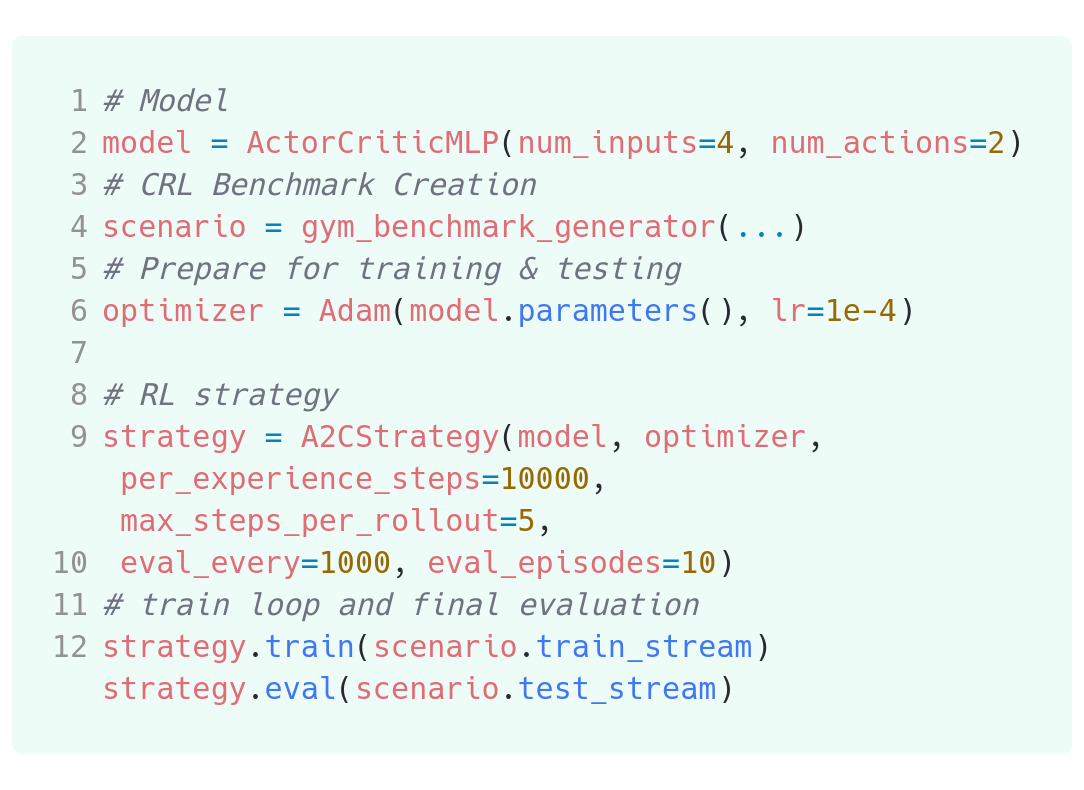}
  \caption{Minimal training setup}
  \label{fig:sub2}
\end{subfigure}
\caption{Example of \texttt{Avalanche RL} usage. (a) defines a task stream alternating two randomly sampled environments for 4 experiences. \texttt{n\_parallel\_envs} specifies the number of parallel actors (Section \ref{vec-env}). The second scenario instead creates a stream of 2 Atari games with pre-processing attached. (b) puts everything together, instantiating a pre-implemented model (Section \ref{extras}) and creating an ``A2C agent" which is trained on the stream of games. The agent will perform 10000 \textit{Update} steps per-experience while gathering 5 data samples at every \textit{Rollout} step (Section \ref{avlrl-train}). Evaluation will take place with the specified parameters.}
\label{fig:benchmarks}
\end{figure}

RL Environments implement a common interface, which is the one of OpenAI Gym environments. This common interface allows to abstract away the interaction with the environment, decoupling the data generation process from the data sampling and freeing the user from the hassle of manually re-writing the data-fetching loop.\\
New CRL benchmarks can be easily created using the \texttt{gym\_benchmark\_generator}, which allows to define an \texttt{RLScenario} by providing any sequence of Gym environments (including custom ones). 
We can see an example in Fig. \ref{fig:benchmarks}, in which we instantiate an \texttt{RLScenario} handling a stream of tasks which gives access to two randomly sampled environments.\\
Note that unlike static datasets, the environment can be used to produce an endless amount of data. Therefore, the interaction with the experience must be explicitly limited by some number of steps or episodes rather than epochs, which we can express during the creation of a \textit{Strategy} as in Section \ref{avlrl-train}.\\\\

As the Atari game suite \cite{atari} has become the main benchmark for RL algorithms in recent years, we also provide a tailored \texttt{atari\_benchmark\_generator} (Fig. \ref{fig:benchmarks}) which takes care of adding common pre-processing techniques (e.g. frame stacking) as Gym Wrappers around each environment. This allows to minimize the time in between experiments as one can easily reproduce setups such as the one in  \cite{ewc} (sampling random Atari games to learn in sequence) by simply specifying a few arguments when creating a scenario. The benchmark interface also promotes the pattern of environment \textit{wrapping}, which is Gym's intended way of organizing data processing methods to favor reproducibility. Reproducibility of experiments in particular is of great importance to \texttt{Avalanche} and one of the main reasons that drove us to propose an end-to-end framework for CRL.

\subsection{Training: Reinforcement Learning Strategies} \label{avlrl-train}
\texttt{Avalanche RL} provides several learning algorithms (listed at the end of this section) which have been implemented to be highly modular and easily customizable. The framework offers full-access to their internals in order to provide fine-grained customization options and specific support for continual learning techniques.\\
There are two main patterns to adapt a learning algorithm: subclassing and \textbf{Plugins} (as introduced in Section \ref{principles}). In particular, \texttt{Avalanche RL} implements most continual learning strategies as Plugins. The modularity of the implementation allows to combine many RL strategies with popular CL strategies, such as replay buffers, regularization methods, and so on. As far as we are aware, \texttt{Avalanche} is the only library that allows the seamless composition of different learning algorithms. RL strategies inherit from \texttt{RLBaseStrategy}, a class which provides a common ``skeleton" for both on and off-policy algorithms and abstracts many of the most repetitive patterns, including environment interaction, tracking metrics, CPU-GPU tensor relocation and more. \texttt{RLBaseStrategy} also provides callbacks which can be used by plugins. \\
Inspired by the open-source framework \texttt{stable-baseline3} \cite{sb3} (sb3), RL strategies are divided into two main steps: \textbf{rollout} collection and \textbf{update}.
Unlike sb3, we grouped both \textit{on} and \textit{off-policy} algorithms under this simple workflow.\\
The rollout stage abstracts the \textit{data gathering} process which iterates the following steps:
\begin{enumerate}
    \item $a_t\sim \pi_{\theta}(s_t)$: \texttt{sample\_rollout\_action}, to be implemented by the specific algorithm, returns the action to perform during a rollout step.
    \item play action and observe next state and reward: \textit{s', r, done, info}\texttt{=env.step($a_t$)} referring to \texttt{Gym} interface.
    \item store state transition in some data structure: \textbf{Step}. Store multiple Steps in a \textbf{Rollout}. These data structures are optimized for insertion speed and lazily delay ``re-shaping" operations until they are needed by the update phase.
    \item test rollout terminal condition, number of steps or episodes to run.
\end{enumerate}

The \textbf{update} step is instead entirely delegated to the specifics of the algorithm: it boils down to implementing a method which has access to the rollouts collected at the previous stage and must define and compute a loss function which is then used to execute the actual parameters update. To enable the user with fine-grained control over the strategy workflow, we added callbacks which are executed just before and after the two stages.\\
At a higher level, the workflow we described happens within a single experience. To learn from a stream, the process is repeated for each experience in the stream, a behavior which is implemented by the \texttt{RLBaseStrategy}. \\
To summarize, one can implement a RL algorithm by sub-classing \texttt{RLBaseStrategy} and implementing the \texttt{sample\_rollout\_action} and \texttt{update} step. For example, A2C can be implemented in less than 30 lines of code \footnote{\url{https://github.com/ContinualAI/avalanche-rl/blob/master/avalanche_rl/training/strategies/actor_critic.py}}. Alternatively, customization of any algorithm is always possible by implementing a plugin, which allows to ``inject" additional behavior, or by subclassing any of the available strategies. All the algorithm implementations expose their internals through class attributes, so one can for instance access the loss externally (e.g. from plugins) simply with \texttt{strategy.loss}. \\
Along with the release of our framework we provide an implementation of A2C and DQN \cite{dqn-paper}, including popular ``variants" with target network \cite{dqn-nature-paper} and DoubleDQN \cite{double-dqn}.

\subsection{Parallel Actors Interaction: VectorizedEnv} \label{vec-env}
Since the data gathering supports any environment exposing the \texttt{Gym} interface, we are also able to automatically parallelize the agent-environment interaction in a transparent way to the user. This common practice \cite{a3c,sb3,salina} relies on using multiple \textit{Actors}, each owning a local copy of the environment in which they perform actions, while synchronizing updates on a shared network; varying the amount of local resources available to each worker we can obtain different \textit{degrees} of \textit{asynchronicity} \cite{impala}, allowing to scale computations on multiple CPUs.

To implement this behavior we leveraged \texttt{Ray} \cite{ray}, a framework for parallel and distributed computing with a focus on AI applications.
\texttt{Ray} abstracts away the parallel (and distributed) execution of code, sharing data between master and workers by serializing \texttt{numpy} arrays, which, in the case of execution on a single machine, are written once to shared memory in read-only mode and only referred to by actors.

This feature is opaque to the user, as it happens entirely inside a \texttt{VectorizedEnv}: this component wraps a single \texttt{Gym} environment and exposes the same interface, while under the hood it instantiates a pool of actors and handles results gathering and synchronization, acting as master.
The API of our implementation was inspired by the work of \texttt{sb3}, although we opted to use \texttt{Ray} as a backend instead of Python's \texttt{multiprocessing} library due to distributed setting support.\\
\texttt{RLBaseStrategy} takes care of wrapping any environment with a \texttt{VectorizedEnv}, so the user can exploit parallel execution by simply specifying the number of workers/environment replicas, as shown in Figure \ref{fig:benchmarks}.

\subsection{Additional Features} \label{extras}
To complement the features we described in the previous sections we also implemented a series of utility components which one expects from a serviceable framework. Most of the changes listed in this section are not as important when taken singularly but as a whole they contribute significantly to \texttt{Avalanche RL} functionalities and as such they are hereby reported.

\begin{itemize}
    \item \textbf{Models} from \textit{seminal} papers such as  \cite{dqn-nature-paper,double-dqn,dqn-paper,a3c} have been re-implemented in Pytorch and are available in the \textit{Models} module.
    \item Evaluation Metrics: \texttt{RLBaseStrategy} automatically records gathered rewards and episode lengths during training, smoothing scalars with a window average by default. 
    Additionally, one can record any significant value (e.g. loss, $\epsilon$-greedy's $\epsilon$) with minimal effort thanks to improved metrics builders. 
    \item \textit{Continual} Control Environments: classic control environments provided by \texttt{Gym} have been wrapped in order to expose hard-coded parameters (e.g. gravity, force..) which can now be modified to obtain varying conditions. This is useful for rapidly testing out algorithms on well renowned problems.   
    \item Extended available \textbf{Plugins}, including EWC \cite{ewc} and a ReplayMemory-based one inspired by works from  \cite{replay-crl} and \cite{selective-replay-crl}.
    \item Miscellaneous tools such as environment wrappers for easily re-mapping actions keys (useful when learning multiple games with a single network) or reducing the action set and an additional \textit{logger} with improved readability. \texttt{Avalanche RL} is compatible with \texttt{Avalanche} logging methods, such as Tensorboard \cite{tensorflow}.
\end{itemize}

\section{Continual Habitat Lab} \label{chl}

\texttt{Continual-Habitat-Lab} (CHL) is a high-level library for FAIR's simulator Habitat \cite{habitat}: inspired by Habitat-Lab, we created a library with the goal of adding support for continual learning. CHL defines the abstraction layer needed to work with a stream of tasks $\{\tau_1, \tau_2..\}$, the core of CL systems.\\
We designed the library to be a shallow wrapper on top of Habitat-Sim functionalities and API while ``steering" its intended usage toward learning applications, enforcing the data generation process to be carried out through online interaction and dropping the need for a pre-computed Dataset of positions altogether.\\
We also revisited the concept of Task to make it simpler and yet give it more control over the environment: while the next-state transition function $p(s' | s, a)$ is implemented by the dynamics of the simulator (Habitat-Sim), we bundled the reward function $r$ into the task definition. To define a Task one must hence define a reward function $r(s, a, s')\rightarrow r$, a goal test function $g(s)\rightarrow \{T,F\}$ and an action space $\mathcal{A}$ as defined by \texttt{Gym}.\\
As Task is meant to be the main component through which the user can inject logic and behavior to be learned by the agent, we give direct access to the simulator at specific times through callbacks (e.g. to change environment condition, lighting, add objects..).\\\\
In order to to natively support CRL a \textbf{TaskIterator} is assigned to the handling of the stream of tasks, hiding away the logic behind task sampling and duration while giving access to the current active task to be used by the environment.\\
We leveraged the multitude of 3D scenes datasets compatible with Habitat-Sim with the goal of specifying changing environment conditions, a most important feature to CL. To do so, we bundled the functionalities regarding scene switch in a sole component named \textbf{SceneManager}. It provides utilities for loading and switching scenes with a few configurable behaviors: scene swapping can happen on task change or after a number of episodes or or actions is reached, even amid a running episode, maintaining current agent configuration and avoiding any expensive simulator re-instantions.\\\\
To offer a easily configurable system we re-designed the configuration system from scratch basing it on the popular \texttt{OmegaConf} library for Python: apart from providing a unified configuration entry-point which can be created programmatically or from a yaml file, the system dynamically maps Task and Actions parameters to configuration options. This allows the user to change experiments conditions by changing class arguments directly from the configuration file.\\\\
Continual Habitat Lab is integrated with Avalanche RL through a specialized benchmark generator (\texttt{habitat\_benchmark\_generator}) that takes care of \textit{synchronizing} the stream of tasks defined in the CHL configuration with the one served to a Strategy. It does so by defining an \textit{experience} each time a task or scene is changed, while serving the same object reference to the Habitat-Sim environment.



\section{Conclusion and Future Work}
In this paper, we have presented two novel libraries for Continual Reinforcement Learning: \texttt{Avalanche RL} and \texttt{Continual Habitat Lab}. We believe that these libraries can be helpful for the CRL community by extending and adapting work from the Continual Learning community on supervised and unsupervised continual learning (\texttt{Avalanche}) while also integrating a realistic simulator (\texttt{Habitat-Sim}) to benchmark CRL algorithms on complex embodied real-life scenarios.\\\\
In particular, \texttt{Avalanche RL} allows users to easily train and evaluate agents on a continual stream of tasks defined as a sequence of any Gym Environment. It is based on implementing a simple API upon the interaction of RL algorithms and task-streams, while offering a fine-grained control over their internals.\\ 
Through \texttt{Avalanche} researchers can exploit and extend the large amount of work done by the Continual Learning community while benefiting from the integration of highly modular and easily extensible RL algorithms.
The library implements a large set of highly desirable features, such as parallel environment interaction, and provides implementations for popular baselines such as EWC \cite{ewc}, including benchmarks, learning strategies and architectures, all of which can be easily instantiated with a single line of code.\\
\texttt{Avalanche RL} can improve code reusability, ease-of-use, modularity and reproducibility of experiments, and we strongly believe that the whole CRL community would benefit from a collective effort such as \texttt{Avalanche RL} as a tool to speed-up the research in the field.\\\\
Having the goal of providing a shared and collaborative open-source codebase for CRL applications, \texttt{Avalanche RL} is constantly looking to add and refine functionalities. In the short term, we plan to implement a broader range of state-of-the art RL algorithms, including (but not limited to) PPO \cite{ppo}, TRPO \cite{trpo} and SAC \cite{sac}. Additionally, we are also looking to increment the number of CL strategies such as \textit{pseudo-rehersal} \cite{pseudo-rehersal,pseudo-rehersal-appl}.\\
We are aiming to keep on expanding the supported simulators targeting a wider range of applications, from robotics to games engines \cite{sc2-cl} to widen the CRL benchmarks suite. 
Finally, we are expecting to merge \texttt{Avalanche RL} into \texttt{Avalanche}, striving to provide a single end-to-end framework for all continual learning applications.
\newpage
\bibliographystyle{splncs04}
\bibliography{citations}
%




\end{document}